\documentclass[10pt, journal] {IEEEtran}
\usepackage{floatrow}
\DeclareFloatFont{normalsize}{\normalsize}
\usepackage{amsmath}
\usepackage{graphicx}%
\usepackage{amsfonts}%
\usepackage{amssymb}
\usepackage{amsthm}  %
\usepackage{subfig}
\usepackage{tabularx}
\usepackage{multirow}
\usepackage{makecell}
\usepackage{adjustbox}
\usepackage{url}
\usepackage{empheq}

\usepackage{caption}

\title{\huge Inertial Sensor Data To Image Encoding For Human Action Recognition}

\author{Zeeshan Ahmad, \textit{Graduate Student Member, IEEE}, Naimul Khan, \textit{Senior Member, IEEE}}

\begin{document}

	\maketitle

\begin{abstract}

Convolutional Neural Networks (CNNs) are successful deep learning models in the field of computer vision. To get the maximum advantage of CNN model for Human Action Recognition (HAR) using inertial sensor data, in this paper, we use four types of spatial domain methods for transforming inertial sensor data to activity images, which are then utilized in a novel fusion framework. These four types of activity images are Signal Images (SI), Gramian Angular Field (GAF) Images, Markov Transition Field (MTF) Images and Recurrence Plot (RP) Images. Furthermore, for creating a multimodal fusion framework and to exploit activity images, we made each type of activity images multimodal by convolving with two spatial domain filters : Prewitt filter and High-boost filter. ResNet-18, a CNN model, is used to learn deep features from multi-modalities. Learned features are extracted from the last pooling layer of each ResNet and then fused by canonical correlation based fusion (CCF) for improving
the accuracy of human action recognition. These highly informative
features are served as input to a multi-class Support Vector Machine (SVM). Experimental results
on three publicly available inertial datasets show the superiority of the proposed
method over the current state-of-the-art.

\end{abstract} 


\section{Introduction}
	
\IEEEPARstart{H}{UMAN} action recognition is a progressive area of research in the field of computer vision and machine learning due to its applications in various spheres of our life such as healthcare~\cite{corbishley2007breathing}, surveillance~\cite{dhiman2019robust} and sports~\cite{zhu2008event}. \let\thefootnote\relax\footnote{An earlier version of this paper is published in 2019 IEEE Fifth International Conference on Multimedia Big Data (BigMM). IEEE, 2019, pp. 429–434. URL: https://ieeexplore.ieee.org/document/8919410}

\let\thefootnote\relax\footnote{© 2021 IEEE. Personal use of this material is permitted. Permission from IEEE must be obtained for all other uses, in any current or future media, including reprinting/republishing this material for advertising or promotional purposes, creating new collective works, for resale or redistribution to servers or lists, or reuse of any copyrighted component of this work in other works.}

Human Action Recognition using inertial data gained significant attention due to the availability of cost-effective smart phones, wearable sensors and their advantages over vision based sensors such as Kinect and RGB cameras. These advantages include the provisioning of data at high sampling rate, ability to work in gloomy and bounded conditions~\cite{ehatisham2019robust}, preserving the privacy of patients while using for healthcare applications, insensitivity to noise and simpler hardware setup.

Earlier or conventional methods for HAR were statistical methods that were based on hand crafted methods of feature extraction such as descriptive statistics~\cite{bao2004activity} and techniques from signal processing~\cite{krause2003unsupervised}. These traditional methods have few
disadvantages: 1) requirement of domain knowledge about
the data and separation of feature extraction and classification parts~\cite{plotz2011feature}; 2) capturing only a subset of the features; resulting in difficulty to generalize for unseen data. The success of deep learning models in the field of computer vision~\cite{ahmad2019humanactionrec} and image classification~\cite{krizhevsky2012imagenet}, especially the CNN, convinced the researchers to use deep learning models for HAR. 

Commonly used inertial sensors for HAR are accelerometer and gyroscope. These sensors provide data in the form of multivariate time series. Existing methods for HAR using inertial sensor data and deep learning models depend upon the utilization of raw multivariate time series using recurrent neural networks (RNN)~\cite{edel2016binarized}, deep belief networks (DBN)~\cite{fang2014recognizing}, restricted Boltzmann machine (RBM)~\cite{hammerla2015pd}, stacked autoencoders~\cite{almaslukh2017effective} and using 1D and 2D CNN~\cite{chen2015deep}. In~\cite{ahmad2019human}, we experimentally proved that converting inertial data from 1D to 2D i.e in the form of images considerably improves the performance of action recognition task. Furthermore, inertial sensor modality has been extensively used with vision sensor modality such as Kinect to provide supplementary information during the fusion of modalities using both statistical~\cite{chen2015improving} and deep learning methods~\cite{dawar2018action}. 

\begin{figure}
	\centering
	\includegraphics[width=1\linewidth]{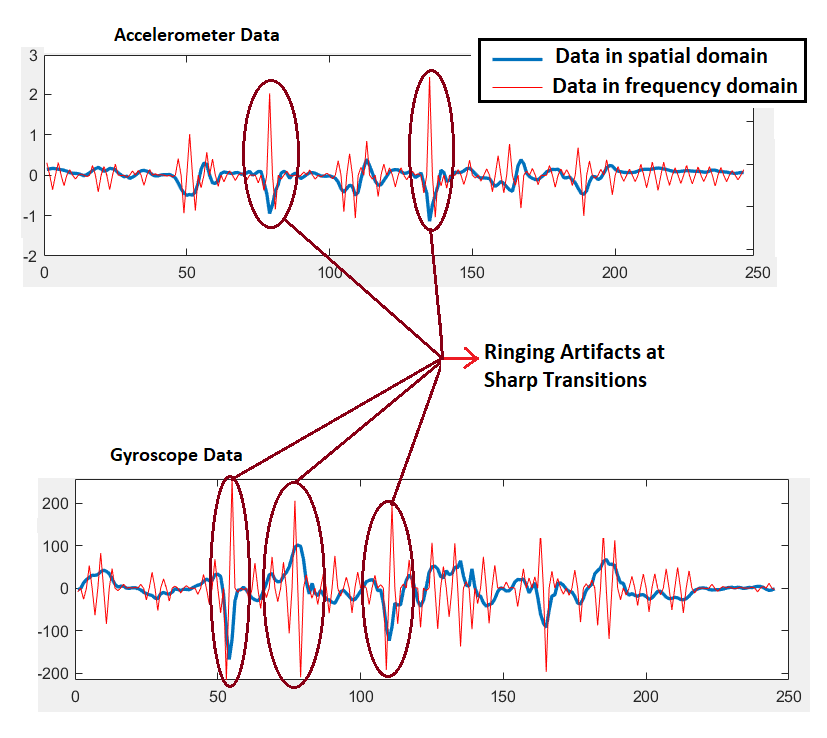}
	\caption{Ringing artifacts are observed at sharp transitions for action ``Bowling'' in UTD-MHAD dataset.}
	\label{fig:Ringing effect}
\end{figure}

In~\cite{jiang2015human}, inertial data for HAR was transformed into 2D activity images for improved performance. Moreover, for further analysis, activity images were transformed to frequecy domain and time-spectrum domain using Discrete Fourier transform (DFT) and 2D Wavelet transform respectively. However, these images in different domains are utilized independently and their fusion could not be exploited. To get the maximum benefit of multidomain fusion for HAR using inertial data, in our previous work~\cite{ahmad2019multidomain}, we proposed CNN
based multidomain multimodal fusion framework that extracts
features in spatial, frequency and time-frequency domain and
hence provide multiscale and multichromatic complementary
features in addition to discriminative features. These multidomain features are then integrated using canonical correlation based
fusion (CCF). Experimental results show that multidomain fusion achieved state-of-the-art performance as compared to previous works. However, we find a deficiency in multidomain fusion. Since inertial sensor data is a time series having sharp transitions or abrupt changes at different instances of its progression, these sharp transitions may not be accurately modeled in frequency domain as sharp transitions in time domain require infinite frequency contents to be modeled in frequency domain. These ringing artifacts are shown in Fig.~\ref{fig:Ringing effect} and we can see overshoots in frequency domain when a sharp transition occurs in spatial domain. These artifacts also affect the transformed domain images and lead to deterioration in accuracy. This can be observed in Table~\ref{tab:Baseline UTDMHAD}.

To improve on our previous work in~\cite{ahmad2019multidomain}, in this paper, we use four spatial domain methods of transforming inertial sensor data for HAR into activity images called Signal Images (SI), Gramian Angular Field (GAF) Images, Markov Transition Field (MTF) Images and Recurrence Plot (RP) Images. Since each kind of inertial sensor data to image encoding shows different patterns and highlights different features of inertial data, therefore the purpose of using four different kind of activity images is to find the best sensor data to image encoding method for HAR. We experimentally observe that GAF and RP images perform better than MTF and signal images. However, GAF images achieve state-of-the-art performance as compared to other three kinds of images for all datasets.

For performing multimodal fusion using the proposed fusion framework shown in Fig.~\ref{fig:Proposed method}, we made each type of activity image multimodal by convolving them with spatial domain Prewitt and high-boost filters as shown in Fig.~\ref{fig:Proposed method}. We train ResNets on each modality. Finally, the learned features extracted from `` pool5 '' layer of each ResNet-18, a CNN model~\cite{he2016deep}, are fused using CCF. Experimental results shown in Table~\ref{tab : Ablation UTDMHAD} prove the dominance of the proposed fusion framework. The proposed method is explained in detail in section~\ref{sec:proposed method} and experimental results are discussed in section~\ref{sec:discussion on results}.

 The key contributions of the presented work are:
\begin{enumerate}	
	
	\item We use four spatial domain methods of transforming raw inertial sensor data into activity images for HAR. To the best of our knowledge, this is the first paper where an inertial sensor data for HAR is transformed into four different kinds of images and a detailed experimental based analysis for each kind of activity image using the proposed fusion framework is presented.
	
	\item Converting raw inertial data into spatial domain activity images and then creation of additional modalities using spatial domain filters preserves the temporal correlation among the samples of time series and thus we achieve better recognition performance
	as compared to the frequency domain transformation methods that cause ringing effects in the images.
		
\end{enumerate}	

The rest of the paper is organized as follows. Section II describes the related works on HAR using inertial sensor data individually and their fusion with other modalities. Section III provides detailed description of the proposed method. In Section IV, we provide detailed experimental analysis, where the aforementioned contributions are analyzed in detail through large number of experiments and comparisons with the state-of-the-art models. Section V concludes the paper.

\section{Related Work}

In this section we present the existing work on HAR using inertial sensor data in spatial domain, transform domain and its fusion with other modalities.  

In~\cite{wang2013position}, an improved algorithm based on Fast Fourier Transform (FFT) is proposed which extracts features from resultant acceleration of the data obtained from smartphones. In~\cite{shah2016encoding}, a frequency-based action descriptor called “FADE” is proposed to represent human actions. FFT is used to transform the signals into frequency domain and then these signals are resampled to exploit the frequency domain features. Finally, Manhattan distance is used for measuring similarities between the actions. In~\cite{chen2015deep}, CNN is used to classify eight
different human actions from the data collected by tri-axial
accelerometer.
In~\cite{jiang2015human}, wearable sensor data is converted into activity images and then CNN is employed for action classification. Experiments on three datasets show that proposed method is robust and highly accurate. In~\cite{tufek2019human}, limited data from accelerometer and gyroscope is used for HAR using CNN,
LSTM and classical machine learning algorithms. Accuracy of action recognition was further increased by applying data augmentation and data balancing techniques. Complex activity recognition system based on cascade classifiers and wearable device data is developed in~\cite{ciabattoni2018complex}. The data was collected by smartphone, smartwatch and bluetooth devices and experimental results show the reliability of the proposed method. In~\cite{xu2020human}, inertial data is converted into GAF images and then a multi-sensor data fusion network called Fusion-MdkResNet is used which can process data collected by
different sensors and fuse data automatically.

In~\cite{abdel2020deep}, a novel deep learning model called multi-scale image encoded HHAR (MS-IE-HHAR) for fine-grained human activity recognition from heterogeneous sensors, is presented. Sensor data is transformed into RGB images and then an important  module of the proposed model called  Hierarchical Multi-scale Extraction (HME) is used for feature extraction. Experimental results show competitive performance of the proposed model. In~\cite{lu2019robust}, accelerometer signals are encoded as color images using a
modified recurrence plot (RP) and then a tiny residual neural network is employed for end-to-end image classification. Experiments show that the proposed framework achieves better performance. In~\cite{qin2020imaging}, mobile sensor data for acceleration and angular velocity is encoded to GAF images and then two deep learning models are used to extract features from GAF images of acceleration and angular velocity. These features are then fused by fusion ResNet for improved performance of human activity recognition. Authors in~\cite{das2020mmhar} presented a new deep learning model called MMHAR-EnsemNet (Multi-Modal Human Activity Recognition Ensemble Network) to perform sensor based human activity using four different modalities. Two separate Convolutional Neural Networks (CNNs) are made for skeleton data while one CNN and one LSTM is trained for RGB images. Accelerometer and Gyroscope data is transformed to signal diagram and then another CNN model is trained on them. Finally, all the outputs of the said models have been used to form an ensemble to improve the performance of HAR. In~\cite{ranieri2020uncovering}, a two-stream ConvNet for action recognition, enhanced with Long Short-Term Memory (LSTM)
and a Temporal Convolution Networks (TCN) is presented  to perform human action recognition. To enhance the recognition process, feature level and late fusions are also investigated using two datasets. In~\cite{hu2020harmonic}, a novel loss function called harmonic loss, which is based on the label replication method to replicate true labels at each sequence step of LSTM models, is presented  to improve the overall classification performance of sensor based HAR. This loss function not only takes all local sequence errors into accounts but also considers the relative importance of different local errors in the training.
Finally, integrated methods based on ensemble learning strategy and harmonic loss are presented and analyzed. In~\cite{zhu2020classification}, the time series radar data for HAR is converted into radar spectrograms using short-time Fourier transform(STFT). Radar spectrogram is treated as a time-sequential vector, and a deep learning model (DL) composed of 1-D convolutional neural networks (1D-CNNs) and recurrent neural networks (RNNs) is proposed. The experiments show that
the proposed DL achieves the highest accuracy than that of existing
2-D CNN methods with fewest number of parameters. Authors in~\cite{wang2020combining} presented a residual and LSTM recurrent
networks-based transportation mode detection algorithm using
multiple light-weight sensors integrated in commodity smartphones. Transportation mode detection, a branch of HAR,
is of great importance in analyzing human travel patterns,
traffic prediction and planning. Residual units are used
to accelerate the learning speed and enhance the accuracy of
transportation mode detection while the aattention models are employed  to learn the significance of different features and
different timesteps to enhance the recognition accuracy. 

A survey article~\cite{wang2018deep} provides a detail about the performance of
current deep learning models and future challenges on sensor
based activity recognition.

Multimodal fusion alleviates the shortcomings of individual modalities and thus improve the performance of HAR. In~\cite{liu2020semantics}, inertial data is transformed into GAF images for performing vision-inertial HAR with RGB videos and then a novel framework, named Semantics-aware
Adaptive Knowledge Distillation Networks (SAKDN) is proposed to enhance
action recognition in vision-sensor modalities by adaptively transferring and distilling the knowledge from multiple
wearable sensors. In~\cite{dawar2018real}, computationally efficient real-time detection and
recognition approach is presented to identify actions in the
smart TV application from continuous action streams using
continuous integration of information obtained from depth and
inertial sensor data. An efficient real-time human action recognition system is developed in~\cite{chen2016real} using decision level fusion of depth and inertial sensor data. Deep learning based decision level multimodal fusion framework is proposed in~\cite{dawar2019data}. CNN is used to extract
features from depth images while Recurrent Neural Netwrok
(RNN) is used to capture features from inertial sensor data.
Data augmentation is also carried out to cop up with limited
size data. Authors in~\cite{hwang2017multi} proposed deep neural network based fusion
of images and inertial data for improving the performance of
human action recognition. Two CNNs were used to extract
features from images and inertial sensors and finally the fused
fearures were used to train an RNN classifier. In~\cite{wei2020simultaneous}, video and inertial
data is converted into 2D and 3D images which are used as
input to 2D and 3D CNN. Finally, decision level fusion is used
to increase the recognition accuracy. In\cite{ehatisham2019robust}, multimodal
feature-level fusion approach for robust human action recognition is presented which utilizes data from multiple sensors, including RGB camera, depth sensor, and wearable inertial sensors. Computationally
efficient features are extracted from the data obtained from RGB-D video camera and inertial body sensors. Experimental results show that feature-level fusion of RGB
and inertial sensors provides best performance for the proposed method.

\begin{figure*}
	\centering
	\includegraphics[width=0.9\linewidth]{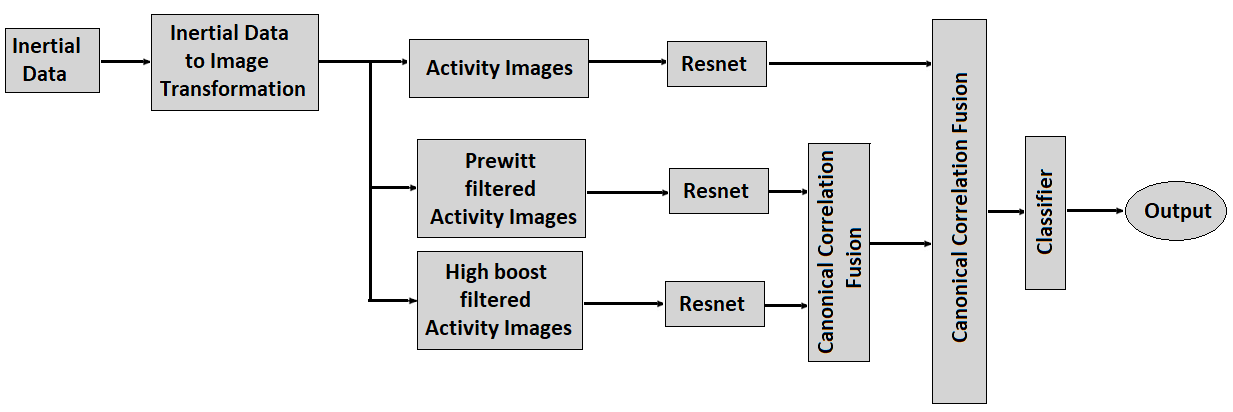}
	\caption{ Complete Overview of Proposed Method}
	\label{fig:Proposed method}
\end{figure*}

To improve our work presented in~\cite{ahmad2019multidomain}, in this paper, we use four spatial domain methods of transforming inertial data for HAR into activity images called Signal Images (SI), Gramian Angular Field (GAF) Images, Markov Transition Field (MTF) Images and Recurrence Plot (RP) Images. For multimodal fusion using the proposed fusion framework shown in Fig.~\ref{fig:Proposed method}, we made each type of activity image multimodal by convolving them with spatial domain Prewitt and high-boost filters as shown in Fig.~\ref{fig:Proposed method}. Finally, the learned features extracted from `` pool5 '' layer of each ResNet are fused using CCF. The proposed method in this paper is an extension of our work in~\cite{ahmad2019multidomain}. Following is the difference between presented work in this paper and our  previous work in~\cite{ahmad2019multidomain}.

\begin{enumerate}
	\item	In our previous work, we transformed inertial sensor data into one kind of activity images only whereas in presented work, we transformed inertial sensor data into four kind of activity images named as Signal Images (SI), Gramian Angular Field (GAF) Images, Markov Transition Field (MTF) Images and Recurrence Plot (RP) Images.
	\item	In previous work, analysis is only performed in transform domain i.e frequency domain and time-spectrum domain using Discrete Fourier transform (DFT) and 2D Wavelet transform respectively whereas in presented work, a detailed experimental based analysis for  both transform domain and spatial domain  is presented. Thus 75\% more analysis is carried out in presented work.
	\item	In previous work, no baseline experiments were performed whereas in presented work, experiments are also performed with two baseline fusion frameworks i.e one baseline in transform domain and other baseline in spatial domain.
	\item	In previous work only quantitative analysis of the results were presented whereas in presented work, both quantitative and qualitative analysis of the results are presented.
	\item The literature review of the presented work is more enriched with recent proposed methods on inertial sensor data for HAR than our previous work.
	
\end{enumerate}
 	
\section{Proposed Method}\label{sec:proposed method}
	
At the input of the proposed fusion framework shown in Fig.~\ref{fig:Proposed method}, we transform the inertial data into activity images. After forming activity images, we made them multimodal by convolving them with Prewitt and high-boost filters. The three modalities i.e activity images, Prewitt filtered activity images and High boost filtered activity images, shown in Fig.~\ref{fig:Proposed method}, are trained simultaneously using three ResNets-18 of same architecture. The learned features are extracted from `` pool5 '' layer of each ResNets-18 and are fused using canonical correlation fusion method. These fused features are finally sent to the SVM classifier for classification task.  The SVM classifier is trained separately from the ResNets-18. ResNets-18 are used as feature extractors, SVM is employed for the final classification. We extract features from `` pool5 '' layer because this is the last layer before classification layer and this layer has more learned features than other layers~\cite{blog2}.  

The block element of proposed model named ``Activity Images", shown in Fig.~\ref{fig:Proposed method}, represents only one kind of activity image at a time. It does not represent the data fusion of four activity images. The only fusion performed in proposed model is the canonical correlation based fusion (CCF) among the activity image and its spatial domain modalities created by Prewitt filter and high boost filter respectively.

 Below we explain each component of the proposed fusion framework.

\subsection{Formation of Activity Images}

Since we transform inertial data into four types of activity images, therefore, we first explain the formation of these images.

\subsubsection{Formation of Signal Images (SI)}

The combination of accelerometer and gyroscope provide multivariate time series data consists of six sequences of signals : three acceleration and three angular velocity sequences.

We converted six sequences into 2D images called signal images based on the algorithm in~\cite{jiang2015human}. Signal image is obtained through row-by-row stacking of given six signal sequences in such a way that each sequence appears alongside to every other sequence.  The signal images are formed by taking advantage of the temporal correlation among the signals. 

Row wise stacking of six sequences has the following order.

\textit{123456135246142536152616}

Where the numbers \textit{1} to \textit{6} represent the sequence numbers in a raw signal. Order of the sequences clearly shows that every sequence neighbors every other sequence to make a signal image. Thus the final width of signal image becomes 24. 

\begin{figure}[h]
	\centering
	\includegraphics[width=0.8\linewidth]{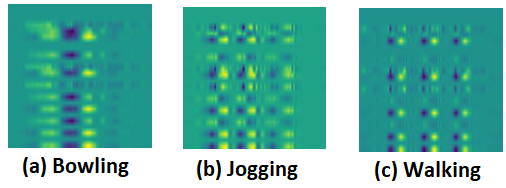}
	\caption{Signal Images of three different actions}
	\label{fig:signal images}
\end{figure}

Length of signal image is decided by making use of sampling rate of datasets which is 50Hz for our two datasets. Therefore, to capture granular motion accurately, the length of the signal image is finalized as 52, resulting in a final image size of 24 x 52. These signal images are shown in Fig.~\ref{fig:signal images} and it is observed that each signal image shows unique pattern and thus enhance the performance of classification. Finally, we resize the images to 224 x 224 for training on resnet.

\subsubsection{Formation of Gramian Angular Field Images (GAF)}\label{sec:gaf image formation}

The image formed by Gramian Angular Field (GAF) represents time series in a polar coordinate system rather than conventional cartesian coordinate system.

Let $X_R\in\mathbb{R}$ be a real valued time series of $n$ samples such that $X_R = \{x_1, x_2, x_3,...,x_l,...,x_n\}$. We rescaled $X_R$ to $X_s$ so that the value of each sample in $X_s$ falls between 0 and 1.
Now we represent the rescaled time series in polar coordinate system by encoding the value as the angular cosine and the time stamp as the radius. This encoding can be understood by the following equation.

\begin{empheq}[right=\empheqrbrace]{equation}
\begin{split}
\phi = arccos(x_{l0}) \\
r = \frac{t_l}{N}
\end{split}
\end{empheq}
where $x_{l0}$ is the rescaled $lth$ sample of the time series, $t_l$ is the time stamp and $N$ is a constant factor to regularize the span of the polar coordinate system.
This transformation has two main advantages. First the transformation is bijective and secondly, it preserves temporal relations through the $r$ cordinates~\cite{wang2015imaging}. From top-left to bottom-right, the image position corresponds to the raw time series is symmetrical
along the main diagonal. Due to this characteristic, the polar coordinates can revert back to the raw
time series~\cite{yang2020sensor}.

The angular perspective of the encoded image can be fully utilized by cosidering the sum/difference between each point to identify the temporal correlation within different time intervals. In this paper we used summation method for Grammian Angular Field and is explained by the following equations

\begin{equation}
GAF = cos(\phi_l + \phi_k)
\end{equation}

\begin{equation}\label{eq : GAF} 
GAF = {X_s}^T . X_s - {\sqrt{I - {X_s}^2}}^T. \sqrt{I - {X_s}^2}
\end{equation}
In equation~\ref{eq : GAF}, $I$ is the unit row vector. Since $n$ is the length of inertial time series data, therefore, the three channel GAF images obtained are of size $n \times n \times 3$. 

\begin{figure}[h]
	\centering
	\includegraphics[width=0.8\linewidth]{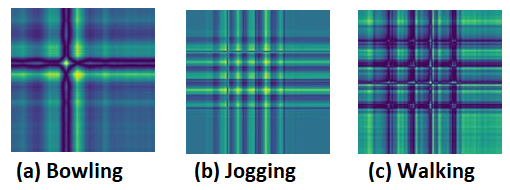}
	\caption{GAF Images of three different actions}
	\label{fig: GAF Image}
\end{figure}

GAF Images obtained from inertial data are shown in Fig~\ref{fig: GAF Image}.

\subsubsection{Formation of Images by Markov Transition Field (MTF)} \label{sec: mtf image formation}

We used the method described in~\cite{wang2015encoding} to encode inertial sensor data into images. Consider the time series $X_R\in\mathbb{R}$ such that $X_R = \{x_1, x_2, x_3,...,x_l,...,x_n\}$. The first step is to identify its $Q$ quantile bins and
assign each $x_l$ to the corresponding bins $q_k (k\epsilon[1, Q])$.
Next step is to construct a $Q \times Q$  weighted adjacency matrix $W$
by counting transitions among quantile bins in the manner of
a first-order Markov chain along the time axis. The normalized form of weighted adjacency matrix is called Markov transition matrix and is unsusceptible to the temporal dependencies, resulting in loss of information. In order to deal with this information loss, Markov transition matrix is converted to Markov transition field matrix (MTF) by spreading out the transition probabilities related to the temporal positions. The Markov transition field matrix is given by 

\begin{equation}\label{eq:Markov transition field matrix}
M=
\begin{bmatrix}
w_{lk|x_1\epsilon q_l,x_1\epsilon q_k}  & \dots & w_{lk|x_1\epsilon q_l,x_n\epsilon q_k}\\
w_{lk|x_2\epsilon q_l,x_1\epsilon q_k} & \dots  & w_{lk|x_2\epsilon q_l,x_n\epsilon q_k}\\
\vdots      &     \ddots &          \vdots    \\
w_{lk|x_n\epsilon q_l,x_1\epsilon q_k}  & \dots & w_{lk|x_n\epsilon q_l,x_n\epsilon q_k}

\end{bmatrix}
\end{equation}
Where $w_{lk}$ is the frequency with which a point in quantile $q_k$ is followed by a point in quantile $q_l$. Since the transformed matrix is formed
by the probabilities of element moving, the MTF method cannot revert to the raw time series data like
GAF. Since MTF requires the time series to be discretized into $Q$ quantile bins to calculate the $Q \times Q$ Markov transition matrix, therefore the size of MTF images is $Q \times Q$.

\begin{figure}[h]
	\centering
	\includegraphics[width=0.8\linewidth]{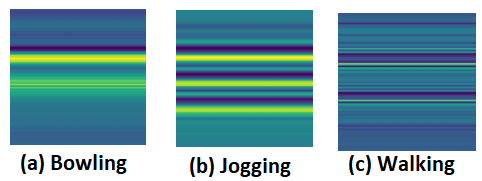}
	\caption{MTF Images of three different actions}
	\label{fig: MTF Image}
\end{figure}

In this Paper, we use 10 bins for the discretization and encoding of inertial data into images. The images formed by MTF is shown in Fig~\ref{fig: MTF Image}.

\subsubsection{Formation of Images by Recurrence Plot (RP)}\label{sec:rp image formation}

Periodicity and irregular cyclicity are the key recurrent behaviors of time series data. The recurrence plots are used as visualization tool for observing the recurrence structure of a time-series~\cite{eckmann1995recurrence}. A recurrence plot (RP) is an image obtained from a multivariate time-series, representing the distances between each time point~\cite{blog}.

Let $q(t)\in\mathbb{R}^{d}$ be a multi-variate time-series. Its recurrence plot is defined as

\begin{equation}\label{ eq:RP}
RP = \theta(\epsilon - ||q(i) - q(j)||)
\end{equation}
In equation~\ref{ eq:RP}, $\epsilon$ is threshold and $\theta$ is called heaviside function. Since $n$ is the length of inertial time series data, therefore, the three channel RP images obtained are of size $n \times n \times 3$.

\begin{figure}[h]
	\centering
	\includegraphics[width=0.8\linewidth]{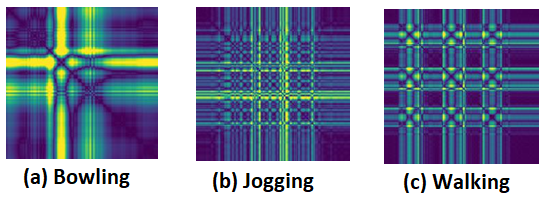}
	\caption{RP Images of three different actions}
	\label{fig: RP Image}
\end{figure}

The images formed by RP is shown in Fig~\ref{fig: RP Image}.

\subsection{Multimodal Fusion}

Since we have converted inertial data into four kinds of activity images. We experiment with each kind of images using our proposed multimodal fusion framework shown in Fig.~\ref{fig:Proposed method}. For experiments with proposed fusion framework, we made each imaging modality multimodal by convolving with Prewitt and High-boost filter.

\subsubsection{Creation of Modality Using Prewitt Filter}

We create a modality for each kind of activity image by convolving images with the Prewitt filter. Prewitt filters are edge detectors and are simple to implement. Creating additional modality using Prewitt filter has been proved significant for improving the performance of action recognition task as additional modality provide complementary features for fusion~\cite{ahmad2019human}.

We apply the following 3-by-3 Prewitt filter for creating a modality.

\begin{equation}
H=
\begin{bmatrix}\label{eq: Prewitt Filter}

1 & 1 & 1 \\
0 & 0 & 0 \\
-1 & -1 & -1

\end{bmatrix}
\end{equation} 

\subsubsection{Creation of Modality Using High-boost Filter}

Another modality is created by convolving each kind of activity images with high-boost filter. High-boost filter assigns proper boosted value to each pixel of image according to its significance~\cite{ahmad2020cnn}. High-boost filter has been used in~\cite{alirezanejad2014effect} for watermark recovery in spatial domain by enhancing the dissimilarity between watermarked and unwatermarked parts of the image. High boost filter is a second order derivative filter obtained by subtracting the low pass filtered version of the image from the scaled input image.

\begin{equation}\label{eq:highboostequation}
f_{hb}(x,y) = Af(x,y) - f_{lp}(x,y)
\end{equation}

where $Af(x,y)$ and $f_{lp}(x,y)$ are the scaled and low pass versions of original image $f(x,y)$ and $A$ is called amplification factor that controls the amount of weight given to the images during convolution.

We apply the following 3-by-3 high-boost filter for creating a modality.

\vspace{0.1cm}
\begin{equation}
K=
\begin{bmatrix}\label{eq:highboostmatrix}

-1 & -1 & -1 \\
-1 & 9 & -1 \\
-1 & -1 & -1

\end{bmatrix}
\end{equation}	

\begin{figure*}
	\centering
	\includegraphics[width=0.8\linewidth]{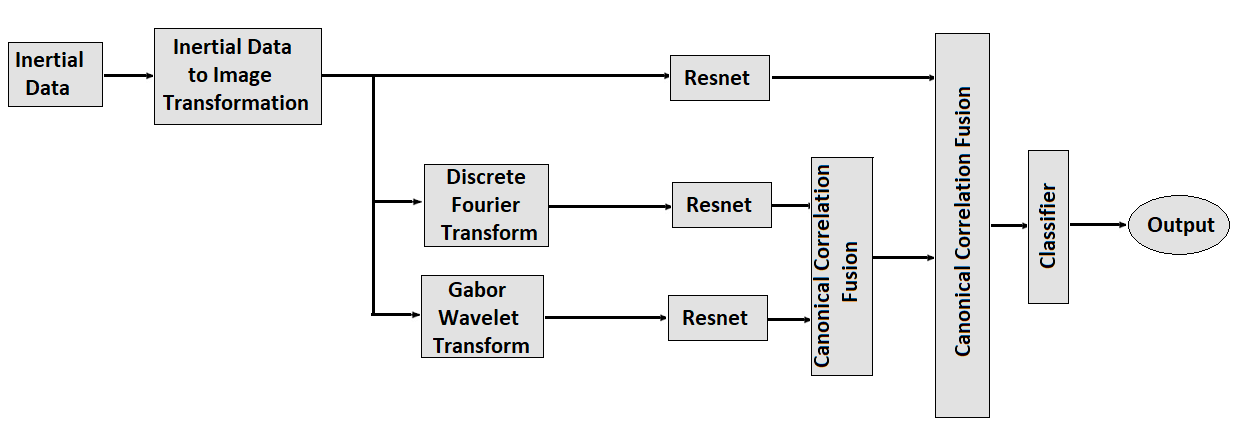}
	\caption{First Baseline Fusion Framework.}
	\label{fig: First Baseline}
\end{figure*}

\begin{figure*}
	\centering
	\includegraphics[width=0.8\linewidth]{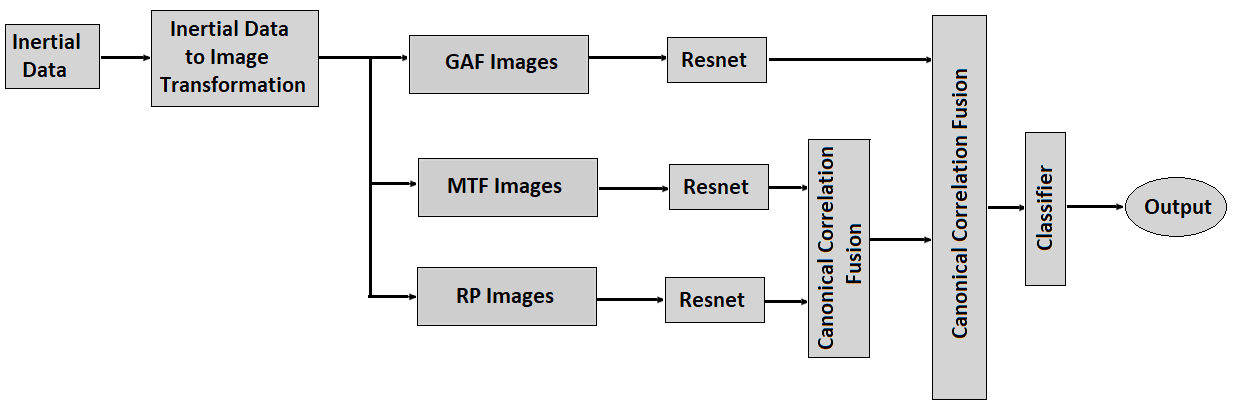}
	\caption{Second Baseline Fusion Framework.}
	\label{fig : Second Baseline}
\end{figure*}

\subsubsection{Canonical Correlation based Fusion (CCF)}\label{CCA Fusion}

After creating modalities, we employ ResNet for extracting features from each modality as shown in Fig.~\ref{fig:Proposed method}. Finally the learned features from `` pool5 '' layer of each ResNet are extracted and fuses in two stages using Canonical correlation based fusion as shown in Fig.~\ref{fig:Proposed method}.

Canonical correlation analysis (CCA) is effective and robust multivariate statistical method for finding the relationship between two sets of variables.

Let $X\in\mathbb{R}^{p \times n}$ and $Y\in\mathbb{R}^{q \times n}$ represents the feature matrices from two modalities, where $p$ and $q$ are the dimensions of the first and second feature set respectively and $n$ are the training samples in each modality. let $\Sigma_{xx}\in \mathbb{R}^{p \times p}$ and $\Sigma_{yy}\in \mathbb{R}^{q \times q}$ denote the within set covariance matrices of $X$ and $Y$ respectively and $\Sigma_{xy}\in \mathbb{R}^{p \times q}$ 
denotes the between set covariance matrix for $X$ and $Y$ and $\Sigma_{yx}=\Sigma_{xy}^T$. The overall augmented covariance matrix of size $(p+q)\times(p+q)$ is given by

\begin{equation}\label{first correlation equation}
cov(X,Y) = 
\begin{bmatrix}

\Sigma_{xx} & \Sigma_{xy} \\
\Sigma_{yx} & \Sigma_{yy}

\end{bmatrix}
\end{equation}

The purpose of CCA is to find the linear combination $X^\prime=AX$ and $Y^\prime=BY$ such that the maximum pairwise correlation between the modalities could be achieved. Matrices $A$ and $B$ are called transformation matrices for $X$ and $Y$ respectively. The correlation between $X^\prime$ and $Y^\prime$ is given by  

\begin{equation}
corr(X^\prime,Y^\prime) = \frac{cov(X^\prime,Y^\prime)}{var(X^\prime).var(Y^\prime)}
\end{equation}
where $cov(X^\prime,Y^\prime)=A^T\Sigma_{xy}B$, $var(X^\prime)= A^T\Sigma_{xx}A$ and $var(Y^\prime)= B^T\Sigma_{yy}B$. $X^\prime$ and $Y^\prime$ are known as canonical variates.

Lagrange's Optimization method is used to maximize the covariance between $X^\prime$ and $Y^\prime$ subject to the constraint that the variance of $X^\prime$ and variance of $Y^\prime$ is equal to unity~\cite{uurtio2018tutorial}.
\[
var(X^\prime)=var(Y^\prime)=1 
\]

On the transformed feature vectors, canonical correlation based fusion (CCF) is performed by  adding the transformed feature vector. Mathematically this addition is written as 

\begin{equation}\label{eq : Fusion equation}
Z =  X^\prime + Y^\prime = A^T X + B^T Y
\end{equation}

The fused features obtained from equation~\ref{eq : Fusion equation} are served as input to a multi-class Support Vector Machine.

We chose SVM classifier over softmax because in our previous work~\cite{ahmad2018towards},  we experimentally showed that SVM performs better than softmax for HAR. Softmax classifier reduces the crossentropy function while SVM employs a margin-based function. Multiclass SVM classifies data by locating the hyperplane at a position where all data points are classified correctly.
Thus, SVM determines the maximum margin among the data points of various classes. The more rigorous nature of classification is likely the reason why SVM performs better than softmax.

\section{Experiments and Results} \label{sec:Experiment and results}

We experiment on three publicly available inertial datasets for human action recognition. These datasets are : Inertial component of UTD Multimodal Human Action Dataset (UTD-MHAD)~\cite{chen2015utd}, Heterogeneity Human Activity Recognition Dataset (HHAR)~\cite{stisen2015smart} and the inertial component of Kinect V2 dataset~\cite{chen2016fusion}. We used subject specific setting for experiments on both datasets by randomly splitting 80\% data into training and 20\% data into testing samples.

We ran the random split 20 times and report the average accuracy and precision. For training on ResNet-18, we resize the images to 224 x 224 using bicubic interpolation. We conduct our experiments on Matlab R2020a on a desktop computer with NVIDIA GTX-1070 GPU. We train ResNet-18 on images till the validation loss stops decreasing further. In addition to this, momentum of 0.9, initial learn rate of 0.005, $L_2$ regularization of 0.004 and mini-batch size of 32 is used to control overfitting. We reached these values through using the grid search method.

\subsection{UTD MHAD Dataset}

This dataset consists of 27 different actions performed by 8 subjects with each subject repeating actions 4 times. 
We only used inertial component of the dataset and convert inertial time series data into activity images as described in section~\ref{sec:proposed method}. We observe that the number of images are less to train ResNet-18. Thus to increase the number of samples we used data augmentation techniques described in~\cite{ahmad2018towards}. 

\subsubsection{Baseline Experiments}

We performed baseline experiments with UTD-MHAD dataset to validate the effect of our proposed fusion framework shown in Fig.\ref{fig:Proposed method}. Two fusion frameworks for baseline experiments are shown in Fig.~\ref{fig: First Baseline} and Fig.~\ref{fig : Second Baseline} respectively. 

The first baseline fusion framework is the same framework proposed in~\cite{ahmad2019multidomain}. In this framework, modalities are created using transform domain transformations such as Discrete Fourier transform and Gabor Wavelet transform respectively. Experimental results in Table~\ref{tab:Baseline UTDMHAD} show that the performance of signal images is worst in transform domains as compared to spatial domain. The performance of signal images in transform domain adversely effect the performance of CCF as shown in Table~\ref{tab:Baseline UTDMHAD}.

The poor performance in transform domain compelled us to redesign the fusion framework. To improve the performance and to create the modalities in spatial domain, we transform inertial data into GAF, MTF and RP images, as described in section~\ref{sec:proposed method}, and design an spatial domain fusion framework shown in Fig.~\ref{fig : Second Baseline}. We used GAF, MTF and RP images in the second baseline fusion framework because we obtained higher accuracy with this combination. With second baseline framework, we obtained an increase in accuracy of 0.8\%.

\begin{table}[h]
	\centering
	\begin{tabular}{c c}
		
		\hline\hline
		\textbf{Methods}& \textbf{Accuracies\%}  \\\hline\hline 
		Signal Images only & 94.2  \\\hline	
		FFT Signal Images & 85.6  \\\hline
		Gabor Signal Images & 87.7\\\hline
		CCA Fusion (Baseline 1) & 96.7\\\hline
		CCA Fusion (Baseline 2) & 97.5 \\\hline
		Decision level Fusion( GAF-MTF-RP) & 95.3  \\\hline
		Inertial data with LSTM~\cite{ahmad2019human} & 86\\\hline
		Inertial data with 1D-CNN~\cite{ahmad2019human} & 77\\\hline\hline
		
	\end{tabular}
	\caption{Baseline Experiments on UTD-MHAD Dataset}
	\label{tab:Baseline UTDMHAD}
\end{table}

\subsubsection{Experiments with Proposed Fusion Framework}

Increased performance of second baseline framework motivated us to further improved the fusion framework. Exemplary performance of Prewitt filter in creating modality in~\cite{ahmad2019human} and high-boost filter in~\cite{ahmad2020cnn} encouraged us to use these filters for creating modalities as shown in Fig.~\ref{fig:Proposed method}. We perform experiments with the proposed fusion framework using all activity images obtained from inertial data. The experimental results in terms of recognition accuracy and precision are shown in Table~\ref{tab : Ablation UTDMHAD}. It is observed that proposed fusion framework beats the baseline experiments convincingly with all kinds of activity images. The accuracies and precisions are calculated using following equations.

\begin{equation}
Accuracy = \frac{TP + TN}{TP + TN + FP + FN}
\end{equation}

\begin{equation}
Precision = \frac{TP}{TP + FP}
\end{equation}

where,

$TP$ = True positive 

$TN$ = True negative

$FP$ = False positive

$FN$ = False negative

Comparison of Accuracies of proposed method with previous methods on inertial component of UTD-MHAD
dataset is shown in Table~\ref{tab:comparison ON UTD-MHAD}.

\begin{table}[h]
	\caption{Ablation Study on UTD-MHAD dataset for Signal Images, GAF Images, MTF Images and RP Images using Proposed fusion framework.}
	\label{tab : Ablation UTDMHAD}
	\centering 
	\begin{adjustbox}{width=1\columnwidth,center}
		\begin{tabular}{c c c c}
			\hline\hline
			\textbf{Image Type}    &\textbf{Methods}  &\textbf{Accuracies\%} &\textbf{Precision\%} \\\hline\hline
			
			\multirow{4}*{\textbf{Signal Images}}
			& {Signal Images only} & {94.1}  & {94.1}  \\\cline{2-4}
			& {Prewitt filtered Signal Images} &{93}  & {93.1}  \\\cline{2-4}
			& {High boost filtered Signal Images} &{95.2} & {95.3}  \\\cline{2-4}
			& {Proposed CCA based Fusion}  &{97.8} & {97.9} \\\hline\hline
			
			\multirow{4}*{\textbf{GAF Images}}
			& {GAF Images only} & {94.7}& {94.8}    \\\cline{2-4}
			& {Prewitt filtered GAF Images} &{94.1}& {94}  \\\cline{2-4}
			& {High boost filtered GAF Images} &{96} & {96.1} \\\cline{2-4}
			& {Proposed CCA based Fusion}  &{98.3}& {98.4} \\\hline\hline
			
			\multirow{4}*{\textbf{MTF Images}}
			& {MTF Images only} & {94.1}& {94}    \\\cline{2-4}
			& {Prewitt filtered MTF Images} &{93.2}& {93.1}  \\\cline{2-4}
			& {High boost filtered MTF Images} &{95.2} & {95.2} \\\cline{2-4}
			& {Proposed CCA based Fusion}  &{98}& {98} \\\hline\hline
			
			\multirow{4}*{\textbf{RP Images}}
			& {RP Images only} & {94.4} & {94.4}   \\\cline{2-4}
			& {Prewitt filtered RP Images} &{94}& {94}  \\\cline{2-4}
			& {High boost filtered RP Images} &{95.8} & {95.9} \\\cline{2-4}
			& {Proposed CCA based Fusion}  &{98.2}& {98.3} \\\hline\hline
			
		\end{tabular}
	\end{adjustbox}
\end{table}

\begin{table}[h]
	\centering
	\begin{tabular}{c c}
		
		\hline\hline
		\textbf{Previous Methods} & \textbf{Accuracy\%}  \\\hline\hline
		
		Ranieri et al.~\cite{ranieri2020uncovering}    &      76 \\\hline
		Das et al.~\cite{das2020mmhar}    &      85.4 \\\hline
		Chen et al.~\cite{chen2016real}      &      88.3 \\\hline
		Ehatisham et al.~\cite{ehatisham2019robust} & 91.6 \\\hline
		Z.Ahmad et al.~\cite{ahmad2018towards}        &      93.7 \\\hline
		Z.Ahmad et al.~\cite{ahmad2019multidomain}        &      95.8 \\\hline
		CCA Fusion (Baseline01) & 96.7\\\hline
		CCA Fusion (Baseline 02) & 97.5 \\\hline
		\textbf{Proposed CCA Fusion (GAF Images)}  & \textbf{98.3} \\
		\hline\hline				
	\end{tabular}
	\caption{ Comparison of Accuracies of proposed method with previous methods on inertial component of UTD-MHAD dataset}
	\label{tab:comparison ON UTD-MHAD}
\end{table}

\subsection{Heterogeneity Human Activity Recognition Dataset}
	
This dataset consists of six actions and is collected through nine users. All users followed a scripted set of activities while carrying eight smartphones (2 instances of LG Nexus 4, Samsung Galaxy
S+ and Samsung Galaxy S3 and S3 mini) and four smart watches (2
instances of LG G and Samsung Galaxy Gear). These smart phones and smart watches are worn by the participents at their waists and arms respectively. Each participant
conducted five minutes of each activity, which ensured a near equal data
distribution among activity classes (for each user and device). We perform same experiments with this dataset, using our proposed fusion framework shown in Fig.~\ref{fig:Proposed method}, as we did with UTD-MHAD. The experimental results are shown in Table~\ref{tab : Ablation HHAR}

\begin{table}[h]
	\caption{Ablation Study on HHAR dataset using Signal Images, GAF Images, MTF Images and RP Images}
	\label{tab : Ablation HHAR}
	\centering 
	\begin{adjustbox}{width=1\columnwidth,center}
		\begin{tabular}{c c c c}
			\hline\hline
			\textbf{Image Type}    &\textbf{Methods}  &\textbf{Accuracies\%} &\textbf{Precision\%}\\\hline\hline
			
			\multirow{4}*{\textbf{Signal Images}}
			& {Signal Images only} & {95.9} & {95.9}    \\\cline{2-4}
			& {Prewitt filtered Signal Images} &{93.8}& {93.8}  \\\cline{2-4}
			& {High boost filtered Signal Images} &{96.7}& {96.7}  \\\cline{2-4}
			& {Proposed CCA based Fusion}  &{98.3}& {98.3} \\\hline\hline
			
			\multirow{4}*{\textbf{GAF Images}}
			& {GAF Images only} & {96.1} & {96.1}   \\\cline{2-4}
			& {Prewitt filtered GAF Images} &{94.9}& {94.9}  \\\cline{2-4}
			& {High boost filtered GAF Images} &{97.1}& {97.1}  \\\cline{2-4}
			& {Proposed CCA based Fusion}  &{99.1}& {99.1} \\\hline\hline
			
			\multirow{4}*{\textbf{MTF Images}}
			& {MTF Images only} & {95.9} & {95.9}   \\\cline{2-4}
			& {Prewitt filtered MTF Images} &{94.1} & {94.1} \\\cline{2-4}
			& {High boost filtered MTF Images} &{96.8}& {96.2}  \\\cline{2-4}
			& {Proposed CCA based Fusion}  &{98.8}& {98.8} \\\hline\hline
			
			\multirow{4}*{\textbf{RP Images}}
			& {RP Images only} & {96.1} & {96.1}   \\\cline{2-4}
			& {Prewitt filtered RP Images} &{94.7}& {94.7}  \\\cline{2-4}
			& {High boost filtered RP Images} &{96.9}& {96.9}  \\\cline{2-4}
			& {Proposed CCA based Fusion}  &{99} & {99} \\\hline\hline
			
		\end{tabular}
	\end{adjustbox}
\end{table}

Comparison of Accuracies of proposed method with previous methods on HHAR
dataset is shown in Table~\ref{tab:comparison on HHAR}.

\begin{table}[h]
	\centering
	\begin{tabular}{c c}
		
		\hline\hline 
		\textbf{Previous Methods} & \textbf{Accuracy\%}  \\\hline\hline 
		
		Yao et al.~\cite{yao2017deepsense}    &      94.5 \\\hline
		Qin et al.~\cite{qin2020imaging}    &      96.8 \\\hline
		Z.Ahmad et al.~\cite{ahmad2019multidomain}        &      98.1 \\\hline
		Abdel-Basset et al.~\cite{abdel2020deep}       &      98.9 \\\hline
		
		\textbf{Proposed CCA Fusion (GAF Images)}  & \textbf{99.1} \\
		\hline\hline				
	\end{tabular}
	\caption{ Comparison of Accuracies of proposed method with previous methods on HHAR dataset}
	\label{tab:comparison on HHAR}
\end{table}

	\subsection{Kinect V2 Dataset}

We use only inertial modality of KinectV2 action dataset. It contains 10 actions performed by six subjects with each subject repeating the action 5 times.  We perform same experiments with this dataset, using our proposed fusion framework shown in Fig.~\ref{fig:Proposed method}, as we did with two previous datasets. The experimental results are shown in Table~\ref{tab : Ablation KinectV2}.

\begin{table}[h]
	\caption{Ablation Study on Kinect-V2 dataset using Signal Images, GAF Images, MTF Images and RP Images}
	\label{tab : Ablation KinectV2}
	\centering 
	\begin{adjustbox}{width=1\columnwidth,center}
		\begin{tabular}{c c c c}
			\hline\hline
			\textbf{Image Type}    &\textbf{Methods}  &\textbf{Accuracies\%}  &\textbf{Precision\%} \\\hline\hline
			
			\multirow{4}*{\textbf{Signal Images}}
			& {Signal Images only} & {96.8} & {96.9}   \\\cline{2-4}
			& {Prewitt filtered Signal Images} &{95.4}& {95.5}  \\\cline{2-4}
			& {High boost filtered Signal Images} &{97.9}& {98}  \\\cline{2-4}
			& {Proposed CCA based Fusion}  &{98.9}& {99} \\\hline\hline
			
			\multirow{4}*{\textbf{GAF Images}}
			& {GAF Images only} & {97.2} & {97.3}   \\\cline{2-4}
			& {Prewitt filtered GAF Images} &{96}& {96.1}  \\\cline{2-4}
			& {High boost filtered GAF Images} &{98.4}& {98.5}  \\\cline{2-4}
			& {Proposed CCA based Fusion}  &{99.5}& {99.5} \\\hline\hline
			
			\multirow{4}*{\textbf{MTF Images}}
			& {MTF Images only} & {96.9} & {96.9}   \\\cline{2-4}
			& {Prewitt filtered MTF Images} &{95.7}& {95.7}  \\\cline{2-4}
			& {High boost filtered MTF Images} &{98.1}& {98.1}  \\\cline{2-4}
			& {Proposed CCA based Fusion}  &{99.2}& {99.2} \\\hline\hline
			
			\multirow{4}*{\textbf{RP Images}}
			& {RP Images only} & {97.1} & {97.2}   \\\cline{2-4}
			& {Prewitt filtered RP Images} &{96}& {96.1}  \\\cline{2-4}
			& {High boost filtered RP Images} &{98.2}& {98.3}  \\\cline{2-4}
			& {Proposed CCA based Fusion}  &{99.4} & {99.5}\\\hline\hline
			
		\end{tabular}
	\end{adjustbox}
\end{table}

\begin{figure}
	\centering
	\includegraphics[width=\linewidth]{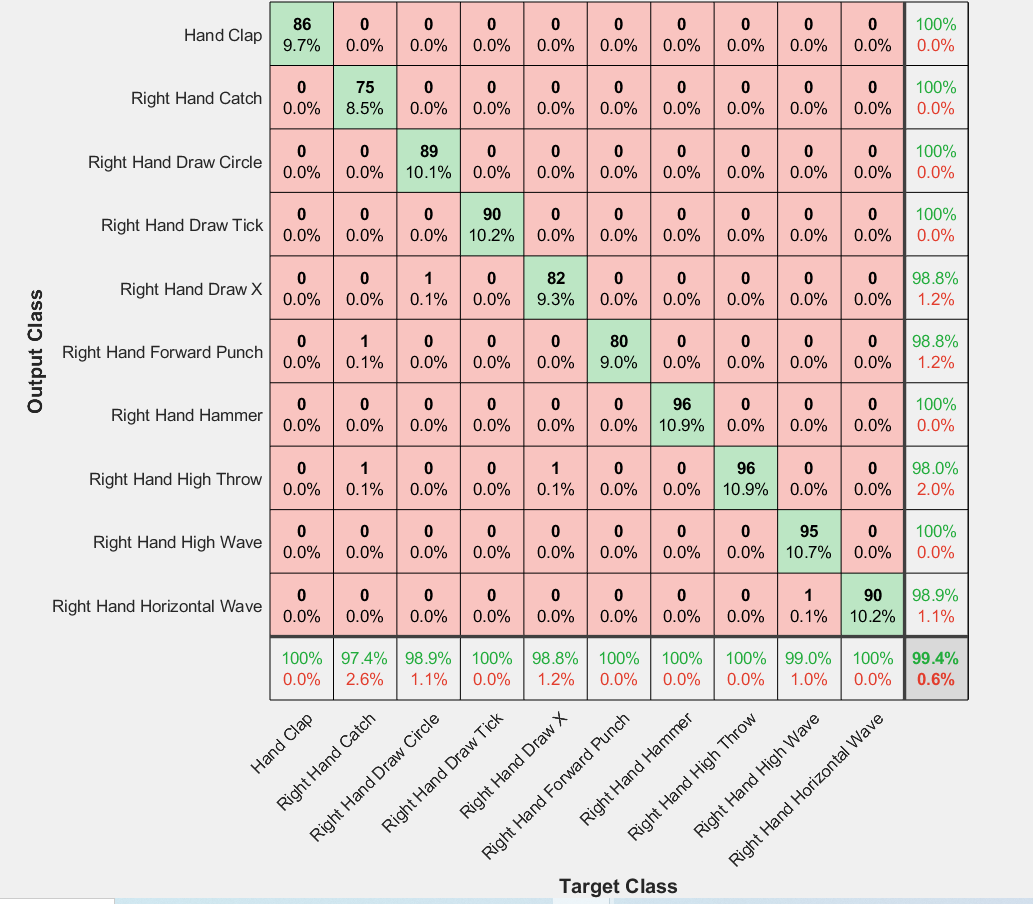}
	\caption{Confusion Matrix of Kinect V2 dataset using proposed method for RP Images.}
	\label{fig : Confmat}
\end{figure}

Comparison of Accuracies of proposed method with previous methods on inertial component of Kinect V2
dataset is shown in Table~\ref{tab:comparison on Kinect V2}. The confusion Matrix of Kinect V2 dataset using proposed method for RP images is shown in Fig.~\ref{fig : Confmat}.

\begin{table}[h]
	\centering
	\begin{tabular}{c c}
		
		\hline\hline
		\textbf{Previous Methods} & \textbf{Accuracy\%}  \\\hline\hline
		
		Chen et al.~\cite{chen2016fusion}     &      96.7 \\\hline
		Z.Ahmad et al.~\cite{ahmad2018towards}        &      96.7 \\\hline
		Z.Ahmad et al.~\cite{ahmad2019multidomain}        &      98.3 \\\hline		
		\textbf{Proposed CCA Fusion (GAF Images)}  & \textbf{99.5} \\
		\hline\hline								
	\end{tabular}
	\caption{ Comparison of Accuracies of proposed method with previous methods on inertial component of Kinect V2 dataset}
	\label{tab:comparison on Kinect V2}
\end{table}

\begin{figure*}
	\centering
	\includegraphics[width=1\linewidth]{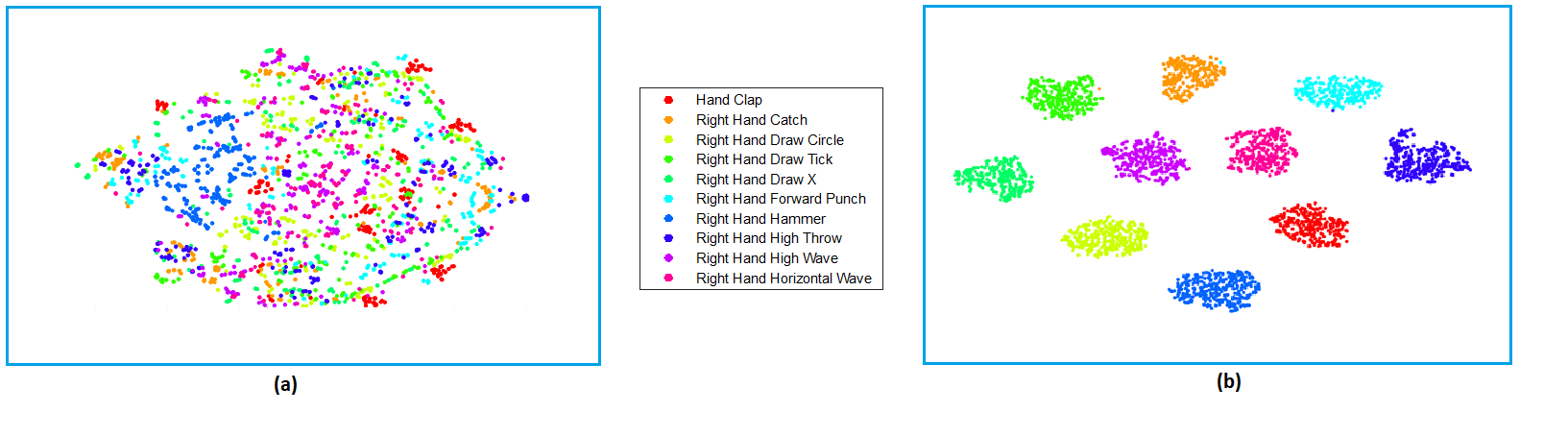}
	\caption{Feature Visualization of Kinect V2 dataset for GAF Images. (a) Features without fusion. (b) Features grouping after CCF using Proposed Method.}
	\label{fig:feature visualization.}
\end{figure*}
	
\subsection{Discussion on Results}\label{sec:discussion on results}

The experimental results of ablation studies, shown in Tables~\ref{tab : Ablation UTDMHAD},~\ref{tab : Ablation HHAR} and~\ref{tab : Ablation KinectV2}, prove that the proposed method perform better than the baseline experiments shown in Table~\ref{tab:Baseline UTDMHAD}.

In the first baseline framework, modalities are created using transform domain methods. These methods could not accurately model the spatial domain abrupt changes in the original data and thus leading to lower accuracy during classification. In second baseline framework, modalities are spatial domain and thus we get improved performance. To build on this increased performance, we design the proposed fusion framework shown in Fig.~\ref{fig:Proposed method} using Prewitt and high-boost filters. During convolution, these filters enhance the most important specifics of the image and thus lead to better performance than the second baseline framework.  

\subsection{Qualitative Analysis of Results}\label{sec: qualitative analysis of results}

For qualitative analysis, we construct feature visualizations of Kinect V2 datset for GAF images using t-SNE~\cite{maaten2008visualizing} as shown in Fig.~\ref{fig:feature visualization.}. In~Fig.~\ref{fig:feature visualization.}(a), features are shown without fusion and we observe that the classes seem to be inseparable. In Fig.~\ref{fig:feature visualization.}(b), we see that features are distinctly separated  and these distinct features improve the performance of a classifier and thus highest recognition accuracy is achieved. Since CCF helps in separating the feature space in a more distinctive manner, we believe that the application of the proposed framework may help improving performance in other application areas as well.

\section{Conclusion}

 In this paper, we used four spatial domain methods of transforming inertial sensor data for Human Action Recognition (HAR) into actvity images. These four types of activity images are Signal Images (SI), Gramian Angular Field (GAF) Images, Markov Transition Field (MTF) Images and Recurrence Plot (RP) Images. We made each type of activity image multimodal by convolving with two spatial domain filters called Prewitt filter and High-boost filter. ResNet-18 are used to extract features from each modality. Learned features from  resnet are extracted and fused by canonical correlation based fusion (CCF) for improving
the accuracy of human action recognition. Experimental results
on three publicly available inertial datasets show the superiority of the proposed
method over the current state-of-the-art.

 \bibliographystyle{IEEEtran}

\begin{thebibliography}{10}
  	\providecommand{\url}[1]{#1}
  	\csname url@samestyle\endcsname
  	\providecommand{\newblock}{\relax}
  	\providecommand{\bibinfo}[2]{#2}
  	\providecommand{\BIBentrySTDinterwordspacing}{\spaceskip=0pt\relax}
  	\providecommand{\BIBentryALTinterwordstretchfactor}{4}
  	\providecommand{\BIBentryALTinterwordspacing}{\spaceskip=\fontdimen2\font plus
  		\BIBentryALTinterwordstretchfactor\fontdimen3\font minus
  		\fontdimen4\font\relax}
  	\providecommand{\BIBforeignlanguage}[2]{{%
  			\expandafter\ifx\csname l@#1\endcsname\relax
  			\typeout{** WARNING: IEEEtran.bst: No hyphenation pattern has been}%
  			\typeout{** loaded for the language `#1'. Using the pattern for}%
  			\typeout{** the default language instead.}%
  			\else
  			\language=\csname l@#1\endcsname
  			\fi
  			#2}}
  	\providecommand{\BIBdecl}{\relax}
  	\BIBdecl
  	
  	\bibitem{corbishley2007breathing}
  	P.~Corbishley and E.~Rodriguez-Villegas, ``Breathing detection: towards a
  	miniaturized, wearable, battery-operated monitoring system,'' \emph{IEEE
  		Transactions on Biomedical Engineering}, vol.~55, no.~1, pp. 196--204, 2007.
  	
  	\bibitem{dhiman2019robust}
  	C.~Dhiman and D.~K. Vishwakarma, ``A robust framework for abnormal human action
  	recognition using {$R$}-{T}ransform and {Z}ernike moments in depth videos,''
  	\emph{IEEE Sensors Journal}, vol.~19, no.~13, pp. 5195--5203, 2019.
  	
  	\bibitem{zhu2008event}
  	G.~Zhu, C.~Xu, Q.~Huang, Y.~Rui, S.~Jiang, W.~Gao, and H.~Yao, ``Event tactic
  	analysis based on broadcast sports video,'' \emph{IEEE Transactions on
  		Multimedia}, vol.~11, no.~1, pp. 49--67, 2008.
  	
  	\bibitem{ehatisham2019robust}
  	M.~Ehatisham-Ul-Haq, A.~Javed, M.~A. Azam, H.~M. Malik, A.~Irtaza, I.~H. Lee,
  	and M.~T. Mahmood, ``Robust human activity recognition using multimodal
  	feature-level fusion,'' \emph{IEEE Access}, vol.~7, pp. 60\,736--60\,751,
  	2019.
  	
  	\bibitem{bao2004activity}
  	L.~Bao and S.~S. Intille, ``Activity recognition from user-annotated
  	acceleration data,'' in \emph{International conference on pervasive
  		computing}.\hskip 1em plus 0.5em minus 0.4em\relax Springer, 2004, pp. 1--17.
  	
  	\bibitem{krause2003unsupervised}
  	A.~Krause, D.~P. Siewiorek, A.~Smailagic, and J.~Farringdon, ``Unsupervised,
  	dynamic identification of physiological and activity context in wearable
  	computing.'' in \emph{ISWC}, vol.~3, 2003, p.~88.
  	
  	\bibitem{plotz2011feature}
  	T.~Pl{\"o}tz, N.~Y. Hammerla, and P.~L. Olivier, ``Feature learning for
  	activity recognition in ubiquitous computing,'' in \emph{Twenty-second
  		international joint conference on artificial intelligence}, 2011.
  	
  	\bibitem{ahmad2019humanactionrec}
  	Z.~Ahmad, K.~Illanko, N.~Khan, and D.~Androutsos, ``Human action recognition
  	using convolutional neural network and depth sensor data,'' in
  	\emph{Proceedings of the 2019 International Conference on Information
  		Technology and Computer Communications}, 2019, pp. 1--5.
  	
  	\bibitem{krizhevsky2012imagenet}
  	A.~Krizhevsky, I.~Sutskever, and G.~E. Hinton, ``Imagenet classification with
  	deep convolutional neural networks,'' in \emph{Advances in neural information
  		processing systems}, 2012, pp. 1097--1105.
  	
  	\bibitem{edel2016binarized}
  	M.~Edel and E.~K{\"o}ppe, ``Binarized-blstm-rnn based human activity
  	recognition,'' in \emph{2016 International conference on indoor positioning
  		and indoor navigation (IPIN)}.\hskip 1em plus 0.5em minus 0.4em\relax IEEE,
  	2016, pp. 1--7.
  	
  	\bibitem{fang2014recognizing}
  	H.~Fang and C.~Hu, ``Recognizing human activity in smart home using deep
  	learning algorithm,'' in \emph{Proceedings of the 33rd chinese control
  		conference}.\hskip 1em plus 0.5em minus 0.4em\relax IEEE, 2014, pp.
  	4716--4720.
  	
  	\bibitem{hammerla2015pd}
  	N.~Hammerla, T.~Ploetz, L.~Rochester, J.~Fisher, R.~Walker, P.~Andras
  	\emph{et~al.}, ``Pd disease state assessment in naturalistic environments
  	using deep learning,'' 2015.
  	
  	\bibitem{almaslukh2017effective}
  	B.~Almaslukh, J.~AlMuhtadi, and A.~Artoli, ``An effective deep autoencoder
  	approach for online smartphone-based human activity recognition,'' \emph{Int.
  		J. Comput. Sci. Netw. Secur}, vol.~17, no.~4, pp. 160--165, 2017.
  	
  	\bibitem{chen2015deep}
  	Y.~Chen and Y.~Xue, ``A deep learning approach to human activity recognition
  	based on single accelerometer,'' in \emph{2015 ieee international conference
  		on systems, man, and cybernetics}.\hskip 1em plus 0.5em minus 0.4em\relax
  	IEEE, 2015, pp. 1488--1492.
  	
  	\bibitem{ahmad2019human}
  	Z.~Ahmad and N.~Khan, ``Human action recognition using deep multilevel
  	multimodal (m2) fusion of depth and inertial sensors,'' \emph{IEEE Sensors
  		Journal}, vol.~20, no.~3, pp. 1445--1455, 2019.
  	
  	\bibitem{chen2015improving}
  	C.~Chen, R.~Jafari, and N.~Kehtarnavaz, ``Improving human action recognition
  	using fusion of depth camera and inertial sensors,'' \emph{IEEE Transactions
  		on Human-Machine Systems}, vol.~45, no.~1, pp. 51--61, 2015.
  	
  	\bibitem{dawar2018action}
  	N.~Dawar and N.~Kehtarnavaz, ``Action detection and recognition in continuous
  	action streams by deep learning-based sensing fusion,'' \emph{IEEE Sensors
  		Journal}, vol.~18, no.~23, pp. 9660--9668, 2018.
  	
  	\bibitem{jiang2015human}
  	W.~Jiang and Z.~Yin, ``Human activity recognition using wearable sensors by
  	deep convolutional neural networks,'' in \emph{Proceedings of the 23rd ACM
  		international conference on Multimedia}.\hskip 1em plus 0.5em minus
  	0.4em\relax Acm, 2015, pp. 1307--1310.
  	
  	\bibitem{ahmad2019multidomain}
  	Z.~Ahmad and N.~M. Khan, ``Multidomain multimodal fusion for human action
  	recognition using inertial sensors,'' in \emph{2019 IEEE Fifth International
  		Conference on Multimedia Big Data (BigMM)}.\hskip 1em plus 0.5em minus
  	0.4em\relax IEEE, 2019, pp. 429--434.
  	
  	\bibitem{he2016deep}
  	K.~He, X.~Zhang, S.~Ren, and J.~Sun, ``Deep residual learning for image
  	recognition,'' in \emph{Proceedings of the IEEE conference on computer vision
  		and pattern recognition}, 2016, pp. 770--778.
  	
  	\bibitem{wang2013position}
  	C.~Wang, J.~Zhang, Z.~Wang, and J.~Wang, ``Position-independent activity
  	recognition model for smartphone based on frequency domain algorithm,'' in
  	\emph{Proceedings of 2013 3rd International Conference on Computer Science
  		and Network Technology}.\hskip 1em plus 0.5em minus 0.4em\relax IEEE, 2013,
  	pp. 396--399.
  	
  	\bibitem{shah2016encoding}
  	D.~Shah, P.~Falco, M.~Saveriano, and D.~Lee, ``Encoding human actions with a
  	frequency domain approach,'' in \emph{2016 IEEE/RSJ International Conference
  		on Intelligent Robots and Systems (IROS)}.\hskip 1em plus 0.5em minus
  	0.4em\relax IEEE, 2016, pp. 5304--5311.
  	
  	\bibitem{tufek2019human}
  	N.~Tufek, M.~Yalcin, M.~Altintas, F.~Kalaoglu, Y.~Li, and S.~K. Bahadir,
  	``Human action recognition using deep learning methods on limited sensory
  	data,'' \emph{IEEE Sensors Journal}, 2019.
  	
  	\bibitem{ciabattoni2018complex}
  	L.~Ciabattoni, G.~Foresi, A.~Monteri{\`u}, D.~P. Pagnotta, L.~Romeo,
  	L.~Spalazzi, and A.~De~Cesare, ``Complex activity recognition system based on
  	cascade classifiers and wearable device data,'' in \emph{2018 IEEE
  		International Conference on Consumer Electronics (ICCE)}.\hskip 1em plus
  	0.5em minus 0.4em\relax IEEE, 2018, pp. 1--2.
  	
  	\bibitem{xu2020human}
  	H.~Xu, J.~Li, H.~Yuan, Q.~Liu, S.~Fan, T.~Li, and X.~Sun, ``Human activity
  	recognition based on gramian angular field and deep convolutional neural
  	network,'' \emph{IEEE Access}, 2020.
  	
  	\bibitem{abdel2020deep}
  	M.~Abdel-Basset, H.~Hawash, V.~Chang, R.~K. Chakrabortty, and M.~Ryan, ``Deep
  	learning for heterogeneous human activity recognition in complex iot
  	applications,'' \emph{IEEE Internet of Things Journal}, 2020.
  	
  	\bibitem{lu2019robust}
  	J.~Lu and K.-Y. Tong, ``Robust single accelerometer-based activity recognition
  	using modified recurrence plot,'' \emph{IEEE Sensors Journal}, vol.~19,
  	no.~15, pp. 6317--6324, 2019.
  	
  	\bibitem{qin2020imaging}
  	Z.~Qin, Y.~Zhang, S.~Meng, Z.~Qin, and K.-K.~R. Choo, ``Imaging and fusing time
  	series for wearable sensor-based human activity recognition,''
  	\emph{Information Fusion}, vol.~53, pp. 80--87, 2020.
  	
  	\bibitem{das2020mmhar}
  	A.~Das, P.~Sil, P.~K. Singh, V.~Bhateja, and R.~Sarkar, ``Mmhar-ensemnet: A
  	multi-modal human activity recognition model,'' \emph{IEEE Sensors Journal},
  	2020.
  	
  	\bibitem{ranieri2020uncovering}
  	C.~M. Ranieri, P.~A. Vargas, and R.~A. Romero, ``Uncovering human multimodal
  	activity recognition with a deep learning approach,'' in \emph{2020
  		International Joint Conference on Neural Networks (IJCNN)}.\hskip 1em plus
  	0.5em minus 0.4em\relax IEEE, 2020, pp. 1--8.
  	
  	\bibitem{hu2020harmonic}
  	Y.~Hu, X.-Q. Zhang, L.~Xu, F.~X. He, Z.~Tian, W.~She, and W.~Liu, ``Harmonic
  	loss function for sensor-based human activity recognition based on lstm
  	recurrent neural networks,'' \emph{IEEE Access}, vol.~8, pp.
  	135\,617--135\,627, 2020.
  	
  	\bibitem{zhu2020classification}
  	J.~Zhu, H.~Chen, and W.~Ye, ``Classification of human activities based on radar
  	signals using 1d-cnn and lstm,'' in \emph{2020 IEEE International Symposium
  		on Circuits and Systems (ISCAS)}.\hskip 1em plus 0.5em minus 0.4em\relax
  	IEEE, 2020, pp. 1--5.
  	
  	\bibitem{wang2020combining}
  	C.~Wang, H.~Luo, F.~Zhao, and Y.~Qin, ``Combining residual and lstm recurrent
  	networks for transportation mode detection using multimodal sensors
  	integrated in smartphones,'' \emph{IEEE Transactions on Intelligent
  		Transportation Systems}, 2020.
  	
  	\bibitem{wang2018deep}
  	J.~Wang, Y.~Chen, S.~Hao, X.~Peng, and L.~Hu, ``Deep learning for sensor-based
  	activity recognition: A survey,'' \emph{Pattern Recognition Letters}, 2018.
  	
  	\bibitem{liu2020semantics}
  	Y.~Liu, G.~Li, and L.~Lin, ``Semantics-aware adaptive knowledge distillation
  	for sensor-to-vision action recognition,'' \emph{arXiv preprint
  		arXiv:2009.00210}, 2020.
  	
  	\bibitem{dawar2018real}
  	N.~Dawar and N.~Kehtarnavaz, ``Real-time continuous detection and recognition
  	of subject-specific smart tv gestures via fusion of depth and inertial
  	sensing,'' \emph{IEEE Access}, vol.~6, pp. 7019--7028, 2018.
  	
  	\bibitem{chen2016real}
  	C.~Chen, R.~Jafari, and N.~Kehtarnavaz, ``A real-time human action recognition
  	system using depth and inertial sensor fusion,'' \emph{IEEE Sensors Journal},
  	vol.~16, no.~3, pp. 773--781, 2016.
  	
  	\bibitem{dawar2019data}
  	N.~Dawar, S.~Ostadabbas, and N.~Kehtarnavaz, ``Data augmentation in deep
  	learning-based fusion of depth and inertial sensing for action recognition,''
  	\emph{IEEE Sensors Letters}, vol.~3, no.~1, pp. 1--4, 2019.
  	
  	\bibitem{hwang2017multi}
  	I.~Hwang, G.~Cha, and S.~Oh, ``Multi-modal human action recognition using deep
  	neural networks fusing image and inertial sensor data,'' in \emph{2017 IEEE
  		International Conference on Multisensor Fusion and Integration for
  		Intelligent Systems (MFI)}.\hskip 1em plus 0.5em minus 0.4em\relax IEEE,
  	2017, pp. 278--283.
  	
  	\bibitem{wei2020simultaneous}
  	H.~Wei and N.~Kehtarnavaz, ``Simultaneous utilization of inertial and video
  	sensing for action detection and recognition in continuous action streams,''
  	\emph{IEEE Sensors Journal}, 2020.
  	
  	\bibitem{blog2}
  	\BIBentryALTinterwordspacing
  	Extracting image features from pretrained netwrok. [Online]. Available:
  	\url{https://www.mathworks.com/help/deeplearning/ug/extract-image-features-using-pretrained-network.html}
  	\BIBentrySTDinterwordspacing
  	
  	\bibitem{wang2015imaging}
  	Z.~Wang and T.~Oates, ``Imaging time-series to improve classification and
  	imputation,'' in \emph{Twenty-Fourth International Joint Conference on
  		Artificial Intelligence}, 2015.
  	
  	\bibitem{yang2020sensor}
  	C.-L. Yang, Z.-X. Chen, and C.-Y. Yang, ``Sensor classification using
  	convolutional neural network by encoding multivariate time series as
  	two-dimensional colored images,'' \emph{Sensors}, vol.~20, no.~1, p. 168,
  	2020.
  	
  	\bibitem{wang2015encoding}
  	Z.~Wang and T.~Oates, ``Encoding time series as images for visual inspection
  	and classification using tiled convolutional neural networks,'' in
  	\emph{Workshops at the Twenty-Ninth AAAI Conference on Artificial
  		Intelligence}, 2015.
  	
  	\bibitem{eckmann1995recurrence}
  	J.~Eckmann, S.~O. Kamphorst, D.~Ruelle \emph{et~al.}, ``Recurrence plots of
  	dynamical systems,'' \emph{World Scientific Series on Nonlinear Science
  		Series A}, vol.~16, pp. 441--446, 1995.
  	
  	\bibitem{blog}
  	\BIBentryALTinterwordspacing
  	Recuplots and cnns for time-series classification. [Online]. Available:
  	\url{https://www.kaggle.com/tigurius/recuplots-and-cnns-for-time-series-classification}
  	\BIBentrySTDinterwordspacing
  	
  	\bibitem{ahmad2020cnn}
  	Z.~Ahmad and N.~Khan, ``Cnn based multistage gated average fusion (mgaf) for
  	human action recognition using depth and inertial sensors,'' \emph{IEEE
  		Sensors Journal}, 2020.
  	
  	\bibitem{alirezanejad2014effect}
  	M.~Alirezanejad, V.~Saffari, S.~Amirgholipour, and A.~M. Sharifi, ``Effect of
  	locations of using high boost filtering on the watermark recovery in spatial
  	domain watermarking,'' \emph{Indian Journal of Science and Technology},
  	vol.~7, no.~4, p. 517, 2014.
  	
  	\bibitem{uurtio2018tutorial}
  	V.~Uurtio, J.~M. Monteiro, J.~Kandola, J.~Shawe-Taylor, D.~Fernandez-Reyes, and
  	J.~Rousu, ``A tutorial on canonical correlation methods,'' \emph{ACM
  		Computing Surveys (CSUR)}, vol.~50, no.~6, p.~95, 2018.
  	
  	\bibitem{ahmad2018towards}
  	Z.~Ahmad and N.~Khan, ``Towards improved human action recognition using
  	convolutional neural networks and multimodal fusion of depth and inertial
  	sensor data,'' in \emph{2018 IEEE International Symposium on Multimedia
  		(ISM)}.\hskip 1em plus 0.5em minus 0.4em\relax IEEE, 2018, pp. 223--230.
  	
  	\bibitem{chen2015utd}
  	C.~Chen, R.~Jafari, and N.~Kehtarnavaz, ``Utd-mhad: A multimodal dataset for
  	human action recognition utilizing a depth camera and a wearable inertial
  	sensor,'' in \emph{2015 IEEE International conference on image processing
  		(ICIP)}.\hskip 1em plus 0.5em minus 0.4em\relax IEEE, 2015, pp. 168--172.
  	
  	\bibitem{stisen2015smart}
  	A.~Stisen, H.~Blunck, S.~Bhattacharya, T.~S. Prentow, M.~B. Kj{\ae}rgaard,
  	A.~Dey, T.~Sonne, and M.~M. Jensen, ``Smart devices are different: Assessing
  	and mitigatingmobile sensing heterogeneities for activity recognition,'' in
  	\emph{Proceedings of the 13th ACM Conference on Embedded Networked Sensor
  		Systems}.\hskip 1em plus 0.5em minus 0.4em\relax ACM, 2015, pp. 127--140.
  	
  	\bibitem{chen2016fusion}
  	C.~Chen, R.~Jafari, and N.~Kehtarnavaz, ``Fusion of depth, skeleton, and
  	inertial data for human action recognition,'' in \emph{2016 IEEE
  		international conference on acoustics, speech and signal processing
  		(ICASSP)}.\hskip 1em plus 0.5em minus 0.4em\relax IEEE, 2016, pp. 2712--2716.
  	
  	\bibitem{yao2017deepsense}
  	S.~Yao, S.~Hu, Y.~Zhao, A.~Zhang, and T.~Abdelzaher, ``Deepsense: A unified
  	deep learning framework for time-series mobile sensing data processing,'' in
  	\emph{Proceedings of the 26th International Conference on World Wide
  		Web}.\hskip 1em plus 0.5em minus 0.4em\relax International World Wide Web
  	Conferences Steering Committee, 2017, pp. 351--360.
  	
  	\bibitem{maaten2008visualizing}
  	L.~v.~d. Maaten and G.~Hinton, ``Visualizing data using t-sne,'' \emph{Journal
  		of machine learning research}, vol.~9, no. Nov, pp. 2579--2605, 2008.
  	
  \end{thebibliography}

  \vspace{-1cm}
  \begin{IEEEbiography}[{\includegraphics[width=0.8in,height=1in,clip]{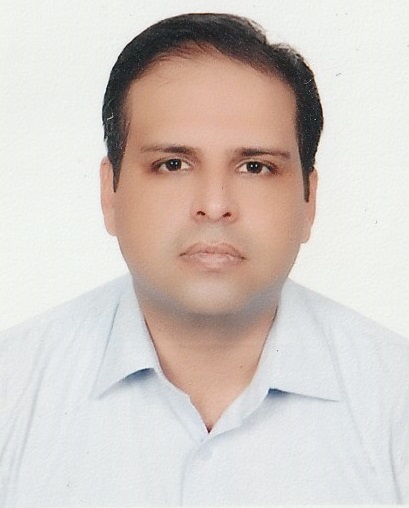}}]{Zeeshan Ahmad} received B.Eng. degree in Electrical Engineering from NED University of Engineering and Technology Karachi, Pakistan in 2001, M.Sc. degree in Electrical Engineering from National University of Sciences and Technology Pakistan in 2005 and MEng. degree in Electrical and Computer Engineering from Ryerson University, Toronto, Canada in 2017.  He is currently pursuing Ph.D. degree with the Department of Electrical and Computer Engineering, Ryerson University, Toronto, Canada. His research interests include Machine learning, Computer vision, Multimodal fusion, signal and image processing.
  \end{IEEEbiography}
\vspace{-1cm}
\begin{IEEEbiography}[{\includegraphics[width=0.8in,height=1in,clip]{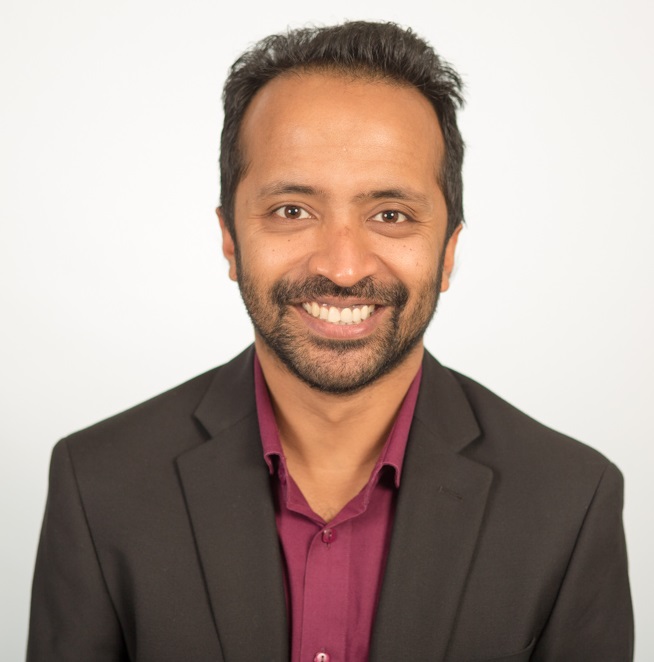}}]{Naimul Khan} is an assistant professor of Electrical and Computer Engineering at Ryerson University, where he co-directs the Ryerson Multimedia Research Laboratory (RML). His research focuses on creating user-centric intelligent systems through the combination of novel machine learning and human-computer interaction mechanisms.  He is a recipient of the best paper award at the IEEE International Symposium on Multimedia, the OCE TalentEdge Postdoctoral Fellowship, and the Ontario Graduate Scholarship. He is a senior member of IEEE and a member of ACM.
\end{IEEEbiography}
  
\end{document}